\documentclass[letterpaper]{article} 
\usepackage{aaai24}  
\usepackage{times}  
\usepackage{helvet}  
\usepackage{courier}  
\usepackage[hyphens]{url}  
\usepackage{graphicx} 
\usepackage{natbib}  
\usepackage{caption} 
\usepackage{algorithm}
\usepackage{algorithmic}

\usepackage{newfloat}
\usepackage{listings}
\DeclareCaptionStyle{ruled}{labelfont=normalfont,labelsep=colon,strut=off} 
\floatstyle{ruled}
\newfloat{listing}{tb}{lst}{}
\floatname{listing}{Listing}
\usepackage{algorithm}
\usepackage{algorithmic}

\usepackage{booktabs}

\usepackage{xcolor,colortbl}
\usepackage{multirow}
\usepackage{comment}

\usepackage{amsmath,amsfonts,bm}

\def\eqref#1{equation~\ref{#1}}

\def\1{\bm{1}}

\DeclareMathAlphabet{\mathsfit}{\encodingdefault}{\sfdefault}{m}{sl}
\SetMathAlphabet{\mathsfit}{bold}{\encodingdefault}{\sfdefault}{bx}{n}

\newcommand{\supp}[1]{#1}
\newcommand{\rot}[1]{}

\def\tedit{t_\mathrm{edit}}

\newcommand{\attmask}{M}

\newcommand{\etal}{\textit{et al}.}

\newcommand{\eg}{\textit{e}.\textit{g}.,}

\newcommand{\tao}[1]{\textcolor{red}{Tao: #1}}

\usepackage{cleveref}

\title{Latent Space Editing in Transformer-Based Flow Matching}
\author {
    Vincent Tao Hu\textsuperscript{\rm 1,2},
    David W Zhang\textsuperscript{\rm 1},
    Pascal Mettes\textsuperscript{\rm 1},
     Meng Tang\textsuperscript{\rm 3},
      Deli Zhao\textsuperscript{\rm 4},
       Cees G.M. Snoek\textsuperscript{\rm 1}
}
\affiliations {
    \textsuperscript{\rm 1} University of Amsterdam,
    \textsuperscript{\rm 2} CompVis Group, LMU Munich,
    \textsuperscript{\rm 3} University of California, Merced,
    \textsuperscript{\rm 4} Alibaba Group\\
}

\begin{document}

\maketitle

\begin{abstract}
    
This paper strives for image editing via generative models. Flow Matching is an emerging generative modeling technique that offers the advantage of simple and efficient training. Simultaneously, a new transformer-based U-ViT has recently been proposed to replace the commonly used UNet for better scalability and performance in generative modeling. Hence, Flow Matching with a transformer backbone offers the potential for scalable and high-quality generative modeling, but their latent structure and editing ability are as of yet unknown. Hence, we adopt this setting and explore how to edit images through latent space manipulation. We introduce an editing space, which we call $u$-space, that can be manipulated in a controllable, accumulative, and composable manner. Additionally, we propose a tailored sampling solution to enable sampling with the more efficient adaptive step-size ODE solvers. Lastly, we put forth a straightforward yet powerful method for achieving fine-grained and nuanced editing using text prompts. Our framework is simple and efficient, all while being highly effective at editing images while preserving the essence of the original content. Our code will be publicly available at \textcolor{cyan}{https://taohu.me/lfm/}

    \end{abstract}

\section{Introduction}

The amazing realism demonstrated by large-scale text-to-image generative models~\cite{rombach2022high_latentdiffusion_ldm,saharia2022photorealistic_imagen,ramesh2021zero_dalle1,ramesh2022hierarchical_dalle2} has garnered significant attention in the research community and beyond, leading to the development of various applications catering to non-expert users. 
For image editing, such models can be directly used in zero-shot fashion~\cite{meng2021sdedit}, enabling users to manipulate images without specific training on editing tasks~\cite{universalguidance}.
The progress made in diffusion models~\cite{ramesh2022hierarchical_dalle2,saharia2022photorealistic_imagen,sgdm} has played a significant role in driving these advancements, prompting further investigations into the understanding of the learned latent space and its potential use for image editing tasks. While current works perform latent space editing on the original latent diffusion model setting and architecture~\cite{hspace_kwon2022diffusion,haas2023discovering}, not much is known about the structure of the latent space in the most recent advances in the field, specifically Flow Matching~\cite{lipman2022flow,liu2022flow,neklyudov2022action,lee2023minimizing} and improved transformer backbones~\cite{uvit}.

Flow Matching~\cite{lipman2022flow} has positioned Continuous Normalizing Flow (CNF) as a strong contender to diffusion models for image synthesis. Flow Matching allows for simulation-free training of CNFs and offers improved efficiency compared to standard diffusion training and sampling techniques. Flow Matching has been quickly integrated in the field, with applications in image generation~\cite{lipman2022flow,sgfm}, video prediction~\cite{davtyan2022randomized}, human motion generation~\cite{motionfm}, point cloud generation~\cite{wu2022fast}, and manifold data~\cite{chen2023riemannian}. 
Orthogonally, recent works have proposed enhancements to the traditional UNet architecture~\cite{unet} used in diffusion models, with the transformer-based U-ViT~\cite{uvit} demonstrating superior scaling performance. Since Flow Matching and U-ViT provide two new pieces of the puzzle towards better generative learning, a natural next step is to discover how images can be manipulated and edited via those techniques.

As a foundation for editing in transformer-based Flow Matching, we seek to uncover whether such an approach has semantic directions that can be manipulated.
We take inspiration from investigations in GANs~\cite{shen2020interfacegan,song2023latent} and diffusion models~\cite{hspace_kwon2022diffusion,jeong2023training,haas2023discovering}, which have revealed there exists a latent space in such networks with semantic directions that can be adjusted and composed. We investigate whether transformer-based Flow Matching also induces a semantic space, that allows us to perform editing in a controllable, accumulative, and composable manner.
%
Furthermore, to fix the misalignment between the forward and backward process in Flow Matching, we propose semantic direction interpolating during the sampling process to reach a more exact and adaptive control over the semantic generation. To make latent space editing viable in practice, manipulation should be done at the input or prompt level. While early work generates new images from scratch based on updated prompts~\cite{rombach2022high_latentdiffusion_ldm,ramesh2022hierarchical_dalle2}, prompt-to-prompt~\cite{p2p} allows for editing images locally while keeping the unedited part similar. Prompt-to-prompt is however specifically designed to work with the cross-attention of U-Net~\cite{unet}. We show that local prompt editing becomes simple and intuitive in U-ViT~\cite{uvit}. Latent space editing becomes as easy as replacing, removing, or appending prompts. Given an initial prompt and generated image, we can simply reweight tokens (e.g. \emph{enlarge beard, remove the dog, shrink tree}), allowing us to locally manipulate images in an invasion-free and user-friendly manner while enabling us to make use of the more powerful transformer architecture.

\begin{figure*}
  \begin{center}
    \includegraphics[width=0.86\textwidth]{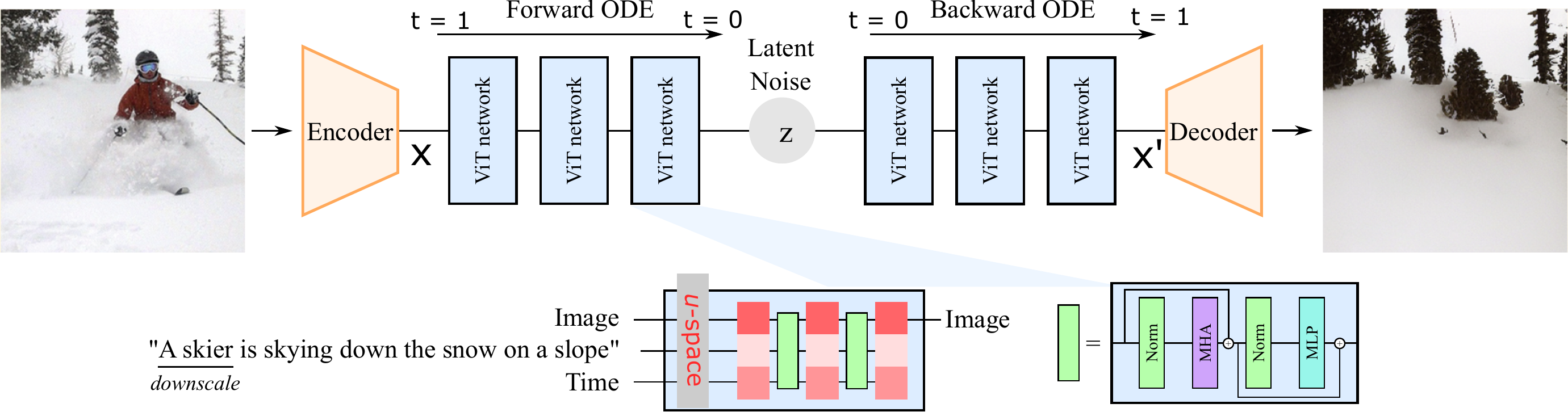}
  \end{center}
  \vspace{-10pt}
  \caption{\textbf{Latent Flow Matching for image editing.} Starting from the original image, we extract the latent feature $x$ from the frozen encoder. Then, Flow Matching is applied in latent space to transfer the trajectory between the latent feature and standard Gaussian noise by integration on the vector field. An editing operation can be triggered in $u$-space  and \texttt{Prompt} by your own desire. The edited feature will be fed back to the decoder to generate the final edited image. 
  }
  \label{fig:lfm}
\end{figure*}

\section{Background: Flow Matching}
\label{sec:flow}

In Flow Matching, we are given a set of samples from an unknown data distribution $q(\mathbf{x})$. The goal is to learn a \textit{flow} that pushes the simple prior density $p_0(\mathbf{x}) {=} {\cal N}(\mathbf{x} \,|\, 0, 1)$ towards a complicated distribution $p_1(\mathbf{x}) {\approx} q(\mathbf{x})$ along the probability path $p_t(\mathbf{x})$. Formally, this is denoted using the push-forward operation as $p_t {=} [\phi_t]_* p_0$.
The time-dependent flow can be constructed via a vector field $\mathbf{v}_t(\mathbf{x}): [0, 1] \times \mathbb{R}^d \rightarrow \mathbb{R}^d$ that defines the flow via the neural ordinary differential equation (ODE):
\begin{align}\label{eq:ode}
    \dot \phi_t(\mathbf{x}) = \mathbf{v}_t(\phi_t(\mathbf{x})), \quad \quad
    \phi_0(\mathbf{x}) &= \mathbf{x}_0 .
\end{align}
Given a predefined probability density path $p_t(\mathbf{x})$ and the corresponding vector field $\mathbf{w}_t(\mathbf{x})$, one can parameterize $\mathbf{v}_t(\mathbf{x})$ with  a neural network with parameter $\theta$  and solve
\begin{align}\label{eq:fm}
    \min_{\theta} \mathbb{E}_{t, p_t(\mathbf{x})} \| \mathbf{v}_t(\mathbf{x};\theta) - \mathbf{w}_t(\mathbf{x}) \|^2.
\end{align}
However, directly optimizing this is infeasible, because we do not have access to $\mathbf{w}_t(\mathbf{x})$ in closed form. Instead, Lipman~\etal~\cite{lipman2022flow} propose to use the conditional vector field $\mathbf{w}_t(\mathbf{x} \,|\, \mathbf{x}_1)$ as the target, which corresponds to the conditional flow $p_t(\mathbf{x} \,|\, \mathbf{x}_1)$. Importantly, they show that this new \textit{conditional Flow Matching} objective 
\begin{align}
    \min_{\theta} \mathbb{E}_{t, p_t(\mathbf{x} \,|\, \mathbf{x}_1), q(\mathbf{x}_1)} \| \mathbf{v}_t(\mathbf{x};\theta) - \mathbf{w}_t(\mathbf{x} \,|\, \mathbf{x}_1) \|^2,
\end{align}
has the same gradients as \Cref{eq:fm}. 
By defining the conditional probability path as a linear interpolation between $p_0$ and $p_1$, all intermediate distributions are Gaussians of the form $p_t(\mathbf{x} \,|\, \mathbf{x}_1) {=} {\cal N}(\mathbf{x} \,|\, tx_1, 1-(1-\sigma_{\text{min}})t)$, where $\sigma_{\text{min}}>0$ is a small amount of noise around the sample $x_1$.
The corresponding target vector field is:
\begin{align}\label{eq:u}
    \mathbf{w}_t(\mathbf{x} \,|\, \mathbf{x}_1) = \frac{\mathbf{x}_1 - (1 - \sigma_{\text{min}}) \mathbf{x}}{1 - (1 - \sigma_{\text{min}})},
\end{align}
Lipman~\etal~\cite{lipman2022flow} show that learning straight paths improves the training and sampling efficiency compared to diffusion paths.
It allows to generate samples by first sampling $\mathbf{x}_0 \sim {\cal N}(\mathbf{x} \,|\, 0, 1)$ and then solving \Cref{eq:ode} using an off-the-shelf numerical ODE solver~\cite{Runge, Kutta, alexander1990solving}. Similarly, we can invert a data point to get its corresponding latent noise by solving the ODE in the other direction.

\section{Latent Space Editing in Flow Matching}

\textbf{Transformer architecture.} Our goal is to facilitate easy-to-use and powerful image editing within the framework of Flow Matching. Transformer architectures not only show promise in enhancing image generation capabilities \cite{uvit,dit_peebles2022scalable}, but they also provide a straightforward editing approach by adjusting the tokens of user prompts. In light of this, we propose a general architecture for transformer-based Flow Matching, which is outlined in \Cref{fig:lfm}. To efficiently generate high-resolution images we use a pretrained autoencoder to compress images $\Tilde{\mathbf{x}}$ into representations $\mathbf{x} {=} E(\Tilde{\mathbf{x}})$ of lower spatial resolution.
While keeping the encoder and decoder parameters fixed, we train a U-ViT~\cite{uvit} using the Flow Matching objective in Equation~\ref{eq:fm}. To generate new images, we first generate a sample $\textbf{x}$ using the CNF and then map it back into the image space using the decoder $\Tilde{\mathbf{x}}' {=} D(\mathbf{x}')$, where $\mathbf{x}' = \texttt{BACKWORD}(\texttt{FORWORD}(\mathbf{x}))$.
In the context of a transformer-based Flow Matching framework, a crucial step to enable image editing through a semantic latent space is to identify a suitable latent space that exhibits semantic properties suitable for manipulation. While a straightforward approach might involve using the middle layer, similar to U-Nets, we have discovered that this approach does not yield a robust semantic latent space. Moreover, it can result in changes that are completely unrelated to the source image or produce images of poor visual quality. We attribute this discrepancy to the absence of a spatial compression in the vision transformer structure, in contrast to the conventional U-Net architecture \cite{unet}. We have found, by experiment, that the most effective space for semantic manipulation lies at the beginning of the U-ViT architecture. To distinguish it from the bottleneck layer, commonly referred to as the $h$-space in the U-Net \cite{hspace_kwon2022diffusion}, we designate this space as $u$-space.


\paragraph{Semantic direction manipulation in $u$-space.}
In order to enable image editing through simple vector additions in the latent space we need to identify the directions in the vector space that correspond to semantically meaningful edits.
We adopt a supervised approach to obtain interpretable semantic directions by contrasting two subset from the dataset: one subset $\{\mathbf{x}_i^{k+}\}_{i}$ consisting of images that possess the desired attribute $k$, and the other subset $\{\mathbf{x}_j^{k-}\}_{j}$ consisting of images without the desired attribute. Examples of attributes include age, gender, or smile.
Importantly, we do not require any paired data, and the images in the two subsets can differ in other arbitrary ways.
We compute a semantic direction $\mathbf{s}^{k}_{t}$ as follows:
\begin{equation}\label{eq:linear_direction}
    \mathbf{s}^{k}_{t} = \frac{1}{n} \sum_{i=1,j=1}^n\left(\mathbf{u}(\mathbf{x}_{i,t}^{k+}) -\mathbf{u}(\mathbf{x}_{j,t}^{k-})\right),
\end{equation}
where $\mathbf{u(\cdot)}$ maps to the semantic latent space, $t$ corresponds to the time variable in~\Cref{eq:ode}, and $i$ and $j$ index the different images in the datasets.

We can collect semantic directions $\mathbf{s}^{k}$ from both real and generated images. In the former case, we gather the latent representations for different time steps through the forward process of the ODE. In the latter case, we utilize the backward process. In this work we focus on semantic directions acquired from real images.
For generated images, we can identify the presence of specific attributes using an off-the-shelf attribute classifier \cite{shen2020interfacegan}.

After collecting the sets of semantic directions, we can proceed to manipulate the sampling (backward) process by editing the $u$-space. This can be achieved through:
\begin{equation}\label{eq:guidance}
    \mathbf{\Tilde{u}}(\mathbf{x}_t, k, w) =  \mathbf{u}(\mathbf{x}_t) + w \cdot  \mathbf{s}^{k}_{t},
\end{equation}
where $w$ is the semantic guidance strength.
%
Image editing can be implemented in $u$-space by replacing $\mathbf{u}(\mathbf{x}_t)$ with $\mathbf{\Tilde{u}}(\mathbf{x}_t, k, w)$ during the forward pass of the neural network.
Additionally, we need to determine the appropriate time step $t$ for performing the injection. Initially, one might consider injecting the guidance signal at every time step when the ODE solver calls the neural network.  However, we have observed that this approach can result in a degradation of the visual quality of the sampled image. To address this issue, we limit the modification to time steps $0<t<t_{edit}$. By focusing the injection on early integration steps, we can improve the visual quality while still achieving effective edits. The hyperparameter $t_{edit}$ can be tuned to find the right balance between the consistency of the edits and the quality of the resulting image.
For a fixed step ODE solver this can be written as:
\begin{equation}
\begin{aligned}
&\mathbf{x}_{t_{i-1}} = 
    &\begin{cases}
    & \text {ODE}(\mathbf{x}_{t_{i}}, \mathbf{\Tilde{u}}(\mathbf{x}_{t_{i}}, k, w)) \quad \text { if } 0 < t_{i} < \tedit \\
    &\text {ODE}(\mathbf{x}_{t_{i}},\mathbf{u}_{t_{i}}) \quad\quad\quad\quad\quad \text { else } 
   
    \end{cases}
\end{aligned}
\label{eq:fixed_step_ode_sample_tedit}
\end{equation}
where $t_{i} = \frac{i}{N}$ and $i \in {0,.... N-1}$, $N$ is the integration number for the ODE solver.

\paragraph{Semantic direction interpolation for adaptive step ODE.} 
Existing methods for editing images using semantic latent directions are limited to fixed step-size ODE solvers.
This is because an adaptive step-size ODE solver requires neural network evaluations at arbitrary time steps $t$, while the semantic directions $\mathbf{s}^{k}_{t}$ were only gathered at fixed intervals. 
This misalignment problem between the time steps keeps previous methods from making use of the more efficient adaptive step-size solvers.
To address this problem, we propose an interpolation-based method to handle the misalignment between the two sequences of time steps.
We first gather the semantic directions using a fixed step-size ODE solver, which only needs to be done once. During the editing process, when the neural network is called for a new time step $t$, we interpolate between the two closest semantic directions in time.
Then we compute the semantic directions during the editing process as:
\begin{align}
    \mathbf{s}_{t}^{k} = \mathbf{s}_{\lfloor {t} \rfloor }^{k}  + (\mathbf{s}_{\lceil {t} \rceil}^{k} - \mathbf{s}_{\lfloor {t} \rfloor}^{k})\times(t - \lfloor {t} \rfloor),
    \label{eq:interp}
\end{align}
where $\lfloor {t} \rfloor$, $\lceil {t} \rceil$ denotes the floor and ceiling values in the grid [$\frac{0}{N}$, $\frac{1}{N}$,..,1]. 
As long as the total number of steps $N$ is large enough, we can assume that the interpolation of the semantic direction can be as accurate as possible. In Algorithm~\ref{alg:attribute_editing} of Appendix, we provide the overall pipeline for semantic direction manipulation in $u$-space with adaptive step-size ODE solvers.

\textbf{Steering by text-conditioned prompts.}
%
In the preceding segment of this section, we focused on image editing through the concept of feature steering, achieved by incorporating semantic offsets in a semantic latent space. Next, we will delve into the realm of text-conditioned generation in Flow Matching, where we aim to explore the potential for users to directly edit images by augmenting text prompts.

First, we revisit the text-conditioned U-ViT architecture~\cite{uvit}. The architecture follows the standard Transformer encoder architecture with additional skip connections similar to a U-Net. The encoded image is patchified into tokens $\phi(x)$, which are then concatenated along the set dimension with the tokens from the text prompt $\psi(\mathcal{P})$ and the embedding of the time step $\mathcal{T}$.

%
%
In every attention layer the tokens  $[\phi(x), \psi(\mathcal{P}), \mathcal{T}]$ are projected to a query matrix $Q = \ell_Q([\phi(x), \psi(\mathcal{P}), \mathcal{T}])$, a key matrix $K = \ell_K([\phi(x), \psi(\mathcal{P}), \mathcal{T}])$, and a value matrix $V = \ell_V([\phi(x), \psi(\mathcal{P}), \mathcal{T}])$, via learned linear projections $\ell_Q, \ell_K, \ell_V$.
The \textit{attention map} is computed as 
\begin{equation}
M=\text{Softmax}\left(\frac{QK^T}{\sqrt{d}}\right),
\end{equation}
where $d$ is the latent projection dimension of the keys and queries. 

In the U-ViT architecture, each attention operation involves interactions between image tokens and the text prompt, which distinguishes it from the U-Net approach where such interactions are restricted to \textit{cross}-attention layers only.
Cross-attention forces every image token to attend to some token in the text prompt even when nothing in the text is relevant to that part of the image. Consequently, modifications to the text prompt can lead to significant image changes that go beyond the intended editing scope. Prompt-to-prompt~\cite{p2p} addresses this issue by augmenting the attention map to retain a similar layout as in the unmodified case. However, this necessitates additional computation and storage of attention maps for the unmodified text prompt version.
On the other hand, in self-attention,  image tokens have the flexibility to choose whether or not to attend to any text token. This limits the impact of modifications made to the text tokens, especially when they are irrelevant to a specific image patch. Motivated by this observation, we explore a simpler form of local prompt editing.

%


In certain scenarios, there is a need for more precise and nuanced editing, such as modifying the magnitude of target concepts, adjusting colors, hairstyles, and so on, instead of making drastic changes like replacing, removing, or adding entire objects.
One straightforward approach is to directly scale the representation of the corresponding token. However, this assumes that the tokens occupy a semantically interpretable space, which may not always hold true. To address this issue, we propose a simpler approach that makes weaker assumptions: scaling the attention value between the modified prompt tokens and image patches.
%
For scaling a specific prompt, we first identify the target token IDs, denoted as $\mathbf{j}^*$. We then apply the following scaling operation:

$$(\attmask^{l}_{t})_{i,j}:=     \begin{cases}
      c \cdot (\attmask^{l}_{t})_{i,j} &\quad\text{if }j\in\mathbf{j}^*\text{and i-th is token from image} \\
        (\attmask^{l}_{t})_{i,j} &\quad\text{otherwise.} \\ 
     \end{cases}$$
     
Here, the parameter $c$ allows for fine-grained and intuitive editing by weakening or strengthening specific parts of the text prompt. The index $l$ denotes the layer of the U-ViT model. In practice, applying this reweighting in every block of the U-ViT architecture yields the best results. \supp{We defer related analysis in Appendix}. 
Similar to editing in the semantic latent space, we find that it is advantageous to limit the modifications to time steps $0 < t < t_{\text{edit}}$.
In our experiments, we observe that this much simpler method can perform a multitude of different editing operations while preserving most of the source image.
    
\section{Experiments}

\begin{figure}
    \centering
    \includegraphics[width=0.5\textwidth]{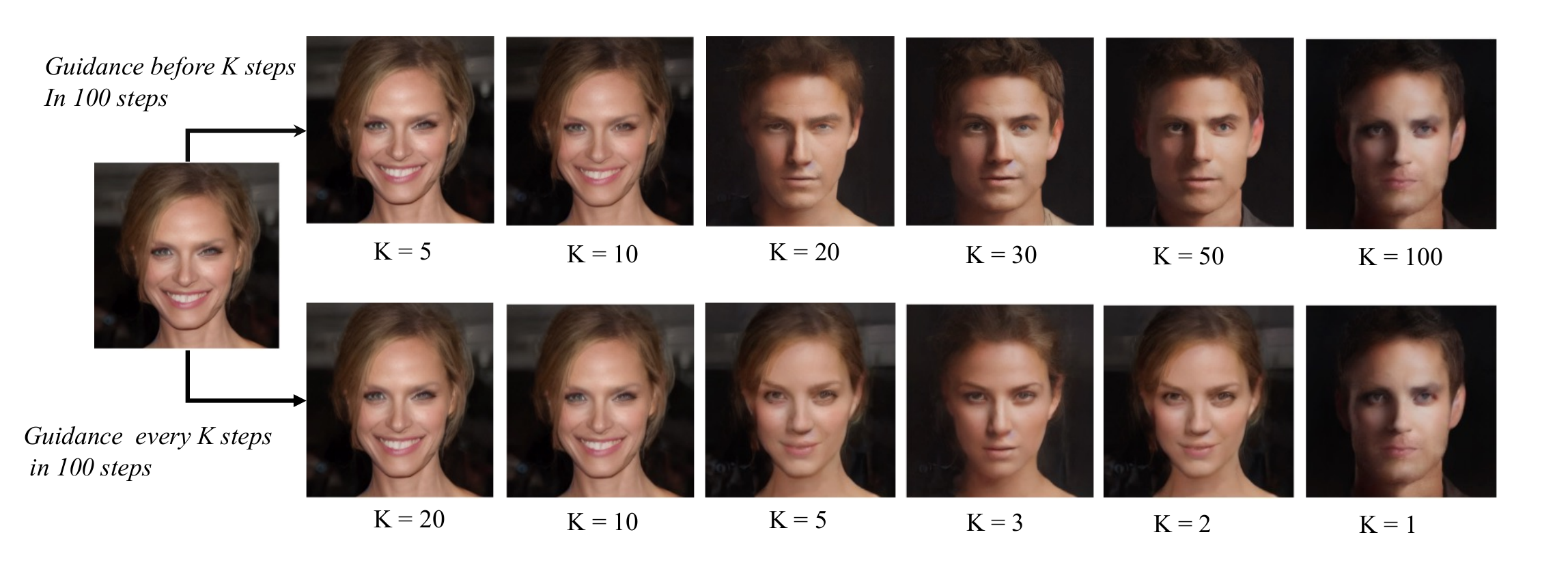}
    \caption{\textbf{The location ablation of the guidance injection}  when editing \textit{male}. Early time step injection is best, injection during full time step will lead to over-constraining.  The first row is signal injection before the first 5, 10, 20, 30, 50, and 100 steps of the backward ODE process. The second row is guidance injection evenly throughout 100 steps. 
    }
    \label{fig:early_has_semantic_meaning_condensed}
\end{figure}

\textbf{Experimental details.}
For the encoder and decoder architecture, we use convolutional VAEs~\cite{kingma2013auto_vae} with pretrained weights following~\cite{rombach2022high_latentdiffusion_ldm}. 
For the experiments on semantic manipulation in the $u$-space, we mainly use the $256\times 256$ CelebA-HQ~\cite{mmcelebahq} dataset. We list all the hyperparameters in the supplementary. For $t_{edit}$, we found $t_{edit}{=}0.5$ works reasonably well. For the guidance strength $w$ in ~\Cref{eq:fixed_step_ode_sample_tedit}, we observe that $w \in (-2, 2)$ generally provides sufficient flexibility while still producing reasonable results. If not mentioned otherwise, we use the adaptive ODE solver \texttt{dopri5}.

For prompt-based editing, we conduct the experiments on the MultiModal-CelebA-HQ~\cite{mmcelebahq} and MS COCO~\cite{coco} datasets, with image resolution $256\times 256$. Both datasets are composed of text-image pairs for training. Typically, there are 5 to 10 captions per image in COCO and MultiModal-CelebA-HQ.  
For editing real images, we choose the images from the validation set of MS COCO.

\textbf{Semantic direction manipulation in $u$-space.} In Figure \ref{fig:early_has_semantic_meaning_condensed}, we investigate the optimal time interval to inject the guidance signal from the semantic direction. Using a fixed step-size ODE solver for a total of 100 steps, we explore signal injection during the first 5, 10, 20, 30, 50, and 100 sampling steps. We observe that injecting the signal for too few steps, for example for 5 or 10 steps, fails to perform the intended edits. Further extending the signal injection period to the first 30 or 50 steps, we find that the semantic direction can be manipulated more noticeably with the same guidance strength. However, if we extend the signal injection until the end it fails to retain anything from the original image. 
We empirically observe that editing within a range of 30 to 50 steps remains a consistent effect across the entire dataset, indicating a minimal burden for hyperparameter tuning.

\begin{figure*}
    \centering
    \includegraphics[width=0.89\textwidth]{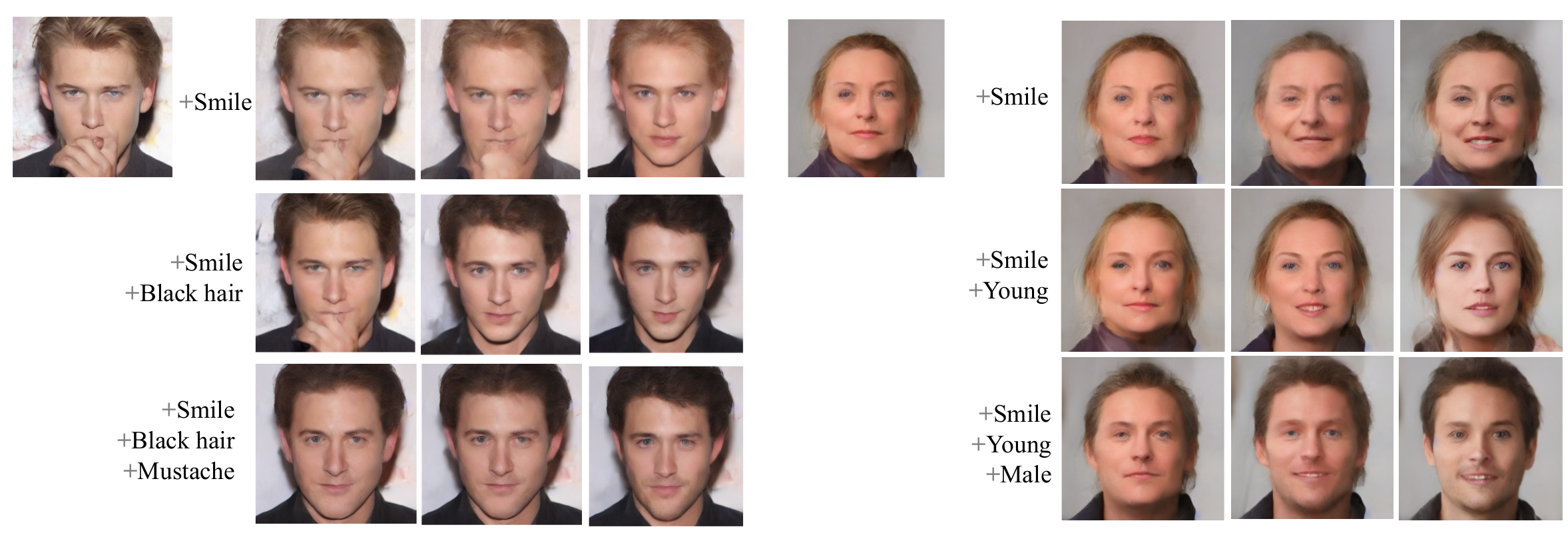}
    \vspace{-10pt}
    \caption{\textbf{Compose multiple attributes sequentially.} We gradually enforce three different semantic directions on a single reference image. The semantic direction of the three attributes is simply averaged by scale $w=2$. 
    }
    \label{fig:attributes_compose}
\end{figure*}

To further validate whether steerability is related to the  time step number, we conducted a controlled experiment in which we injected the guidance signal evenly throughout 100 steps, specifically at every (20, 10, 5, 3, 2, 1) time step. We find that splitting the injection step into those parts did not easily achieve semantic manipulation, even if we manipulated every 5th step into 100, which significantly indicates the importance of the early timestep involvement. \supp{We refer the reader to the progressive visualization in Appendix for more information. }

Furthermore, in Figure~\ref{fig:fm_manipulate_demo} of Appendix, we demonstrate attribute editing on real and sampled images. Our method can achieve various manipulations, such as adding a \textit{smile} or changing the gender to \textit{male}. This not only indicates the effectiveness of the $u$-space but also demonstrates the flexibility of our method on both real and sampled images.

In the end, we validate the compositional ability of these semantic directions in~\Cref{fig:attributes_compose}. We progressively apply the semantic injections one after another. We use the \texttt{euler} solver with 100 steps for the forward process, and \texttt{dopri5} solver for the backward process. By gradually adding the attributes of \textit{smile}, \textit{young}, and \textit{male}, we demonstrate that the sampling process in $u$-space can be manipulated in an accumulative and composable manner.

\begin{figure}
    \centering
    \includegraphics[width=0.5\textwidth]{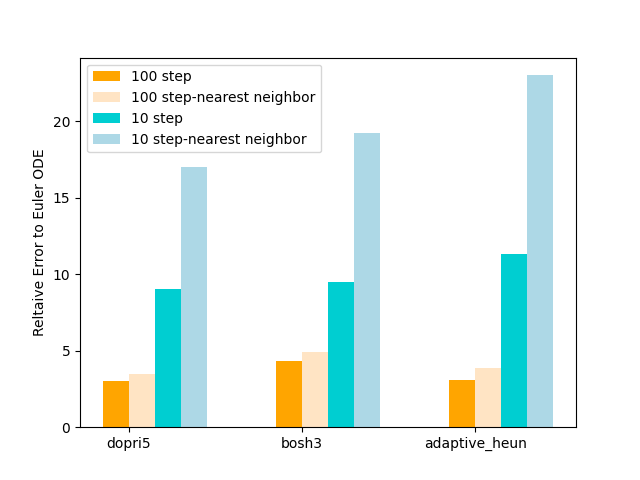}
    \vspace{-28pt}
    \caption{\textbf{Relative error comparison} between adaptive and \texttt{euler} ODE solvers with different time steps.}
    \label{fig:delta_change_bar}
\end{figure}

\begin{figure}
    \centering
    \includegraphics[width=0.5\textwidth]{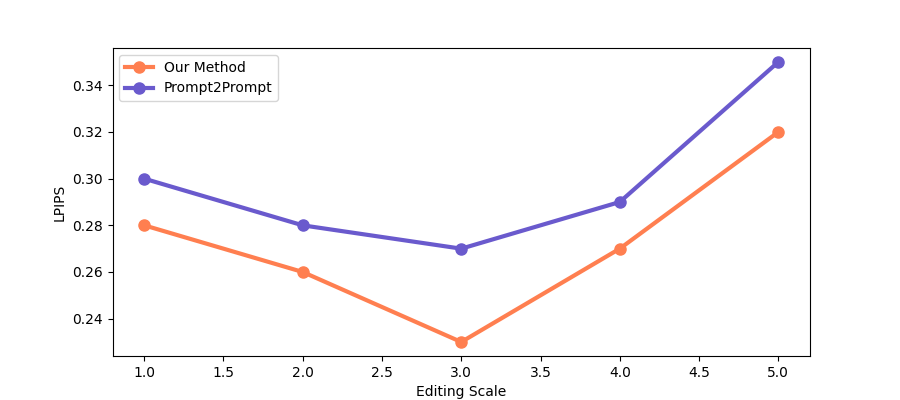}
    \vspace{-20pt}
    \caption{\textbf{Comparison of LPIPS scores with prompt-to-prompt~\cite{p2p} when changing the reweighting scale from 1 to 5.}}
    \label{fig:uspace_lpips}
\end{figure}

\begin{figure*}
  \begin{center}
    \includegraphics[width=0.88\textwidth]{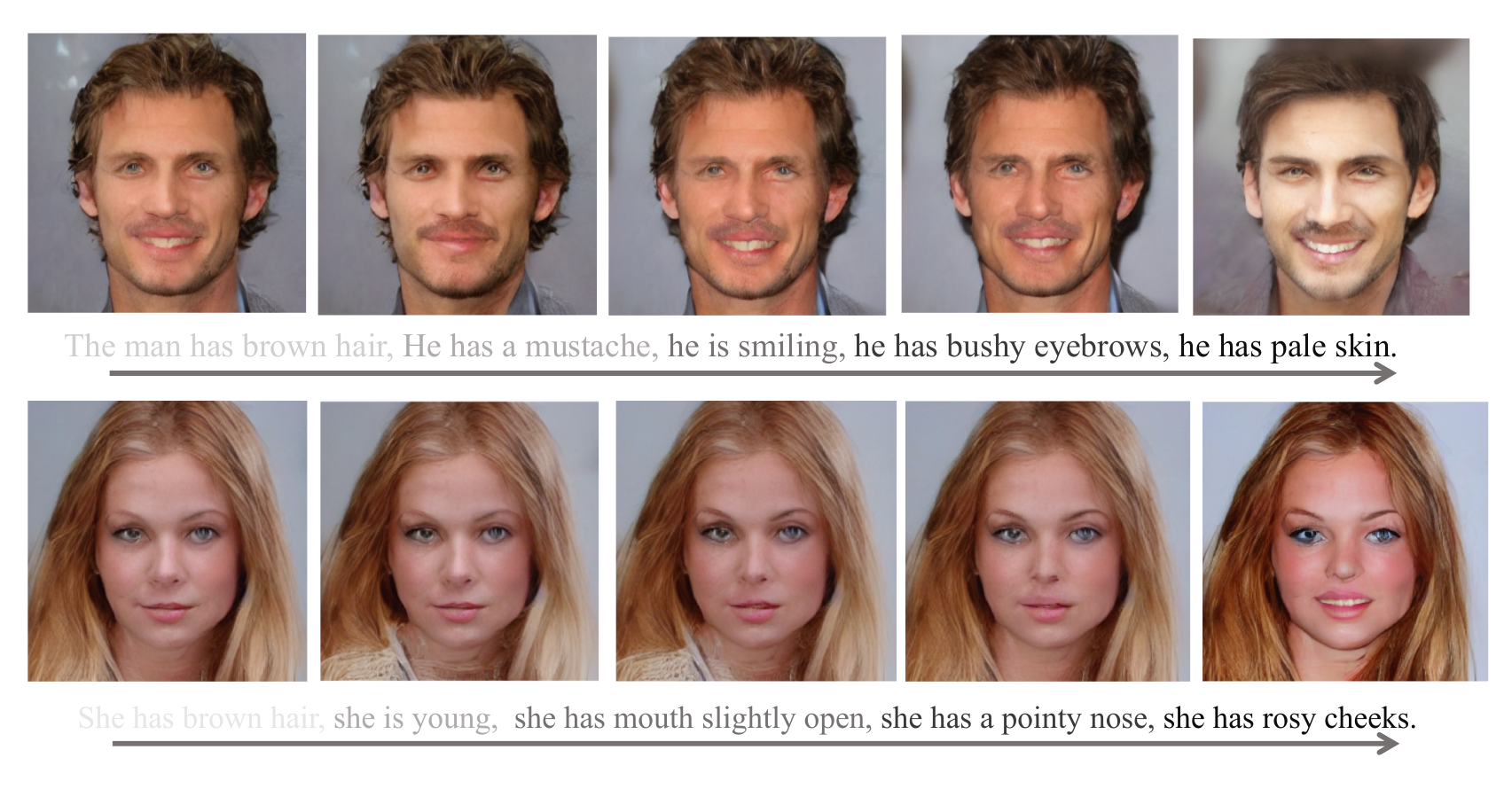}
  \end{center}
  \vspace{-20pt}
  \caption{\textbf{Appending new prompts sequentially}. We can inject semantic meaning directly into the subject while keeping it almost unchanged. To do so, we start by sampling the first image based on the initial prompt. Then, we append the remaining prompts one by one. }
 \label{fig:fm_t2i_vit_sequence}
\end{figure*}

\paragraph{Semantic direction interpolation error analysis.} 
In this study, we investigate the amount of error that may be introduced using semantic direction interpolation. We examine two factors: the number of time steps ($N$) and different ODE solvers, including \texttt{dopri5}~\cite{dormand1980family}, \texttt{bosh3}, and \texttt{adaptive\_heun}. We compare the relative edit error to the editing performed by the \texttt{euler} solver with $N=100$, using guidance strength $w=1$ for the \textit{male} attribute. Additionally, we compare our results to a baseline of nearest neighbor seeking in the grid of $[0,\frac{1}{N},\frac{2}{N}...,1]$ for the semantic direction. In Figure~\ref{fig:delta_change_bar}, we observe that larger time steps ($N$) lead to smaller errors. The relative error is generally at the same level across all three solvers. However, too small of a time step, such as $N=10$, can result in a drastically sizeable relative error. This indicates that using a too-small time step number(large step size) will lead to inaccurate integration for both methods. However, when the time step is large enough, the integration will be more stable and accurate.  
%

\textbf{Text-to-image editing.}
\label{exp:localprompt}
We first demonstrate the accumulation ability of our local-prompt method.  In \Cref{fig:fm_t2i_vit_sequence}, we demonstrate that by appending prompts sequentially based on the initial prompt, the sampled image retains its identity while gaining attributes that align well with the newly added prompts. To the best of our knowledge, a similar experiment has not been conducted in prompt-to-prompt~\cite{p2p}. This suggests that our method can inject semantic meaning by altering raw prompts and that this injected semantic concept can be accumulated without compromising the identity.

\begin{figure}
  \begin{center}
    \includegraphics[width=0.5\textwidth]{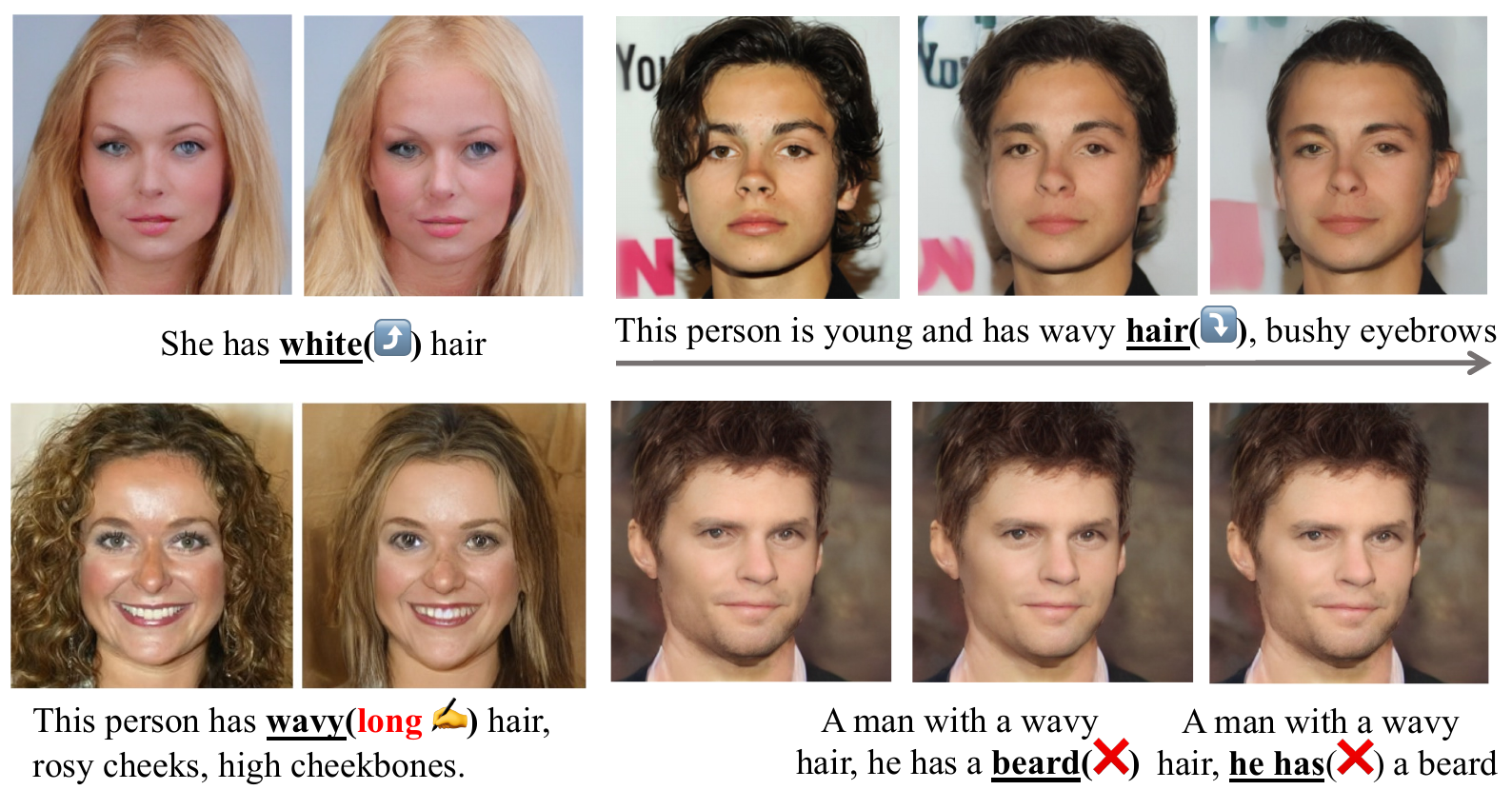}
  \end{center}
  \vspace{-10pt}
  \caption{\textbf{Prompt reweighting, removing, replacing in MM-CelebA-HQ dataset}. Our method can gradually scale the target prompt, remove the target prompt, and be robust to noise during prompt removal. }
 \label{fig:fm_t2i_vit_steering}
\end{figure}

Secondly, we investigate the effects of prompt replacement and removal, as depicted in Figure \ref{fig:fm_t2i_vit_steering}. Notably, we observe that concepts such as \textit{wavy} and \textit{beard} can undergo substantial alterations or even be entirely removed from the resulting image. Intriguingly, when less informative prompts like \textit{he has} are removed, we find that the sampled image remains unchanged, thereby demonstrating the robustness of our method.

\begin{figure*}
  \begin{center}
    \includegraphics[width=1.0\textwidth]{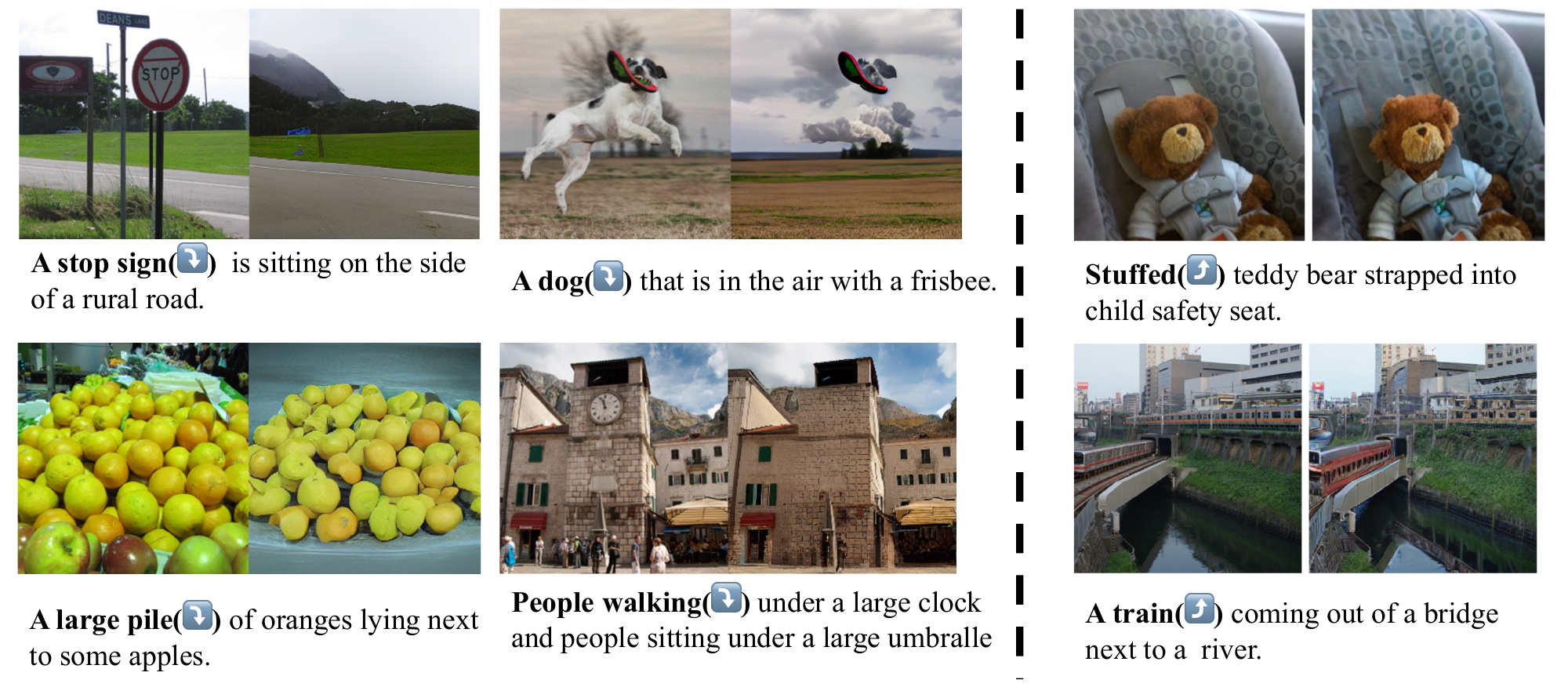}
  \end{center}
  \vspace{-20pt}
  \caption{\textbf{Prompt reweighting} on the real images of  MS COCO dataset by weighting down (left) and weighting up (right). }
 \label{fig:coco_scaledown_up}
\end{figure*}

Lastly, we explore prompt reweighting. Specifically, we perform prompt reweighting on sampled images from MM-CelebA-HQ, as shown in \Cref{fig:fm_t2i_vit_steering}. By reweighting the attributes of \textit{white} and \textit{hair}, we can modify them effectively without altering the identity. In the top right row of \Cref{fig:fm_t2i_vit_steering}, we demonstrate a progressive and smooth evolution by changing the reweighting scale $w$ from $2$ to $-10$. The smooth change indicates a well-aligned embedding between text and image space.
To explore the reweighting ability in a more complex dataset, we conduct experiments on MS COCO, shown in~\Cref{fig:coco_scaledown_up}. We used a pretrained latent flow-matching model on MS COCO and were able to easily invert the image from $\mathbf{x}_{1}$ to latent $\mathbf{x}_{0}$. We progressively conduct attention map editing as we revert the latent back from $\mathbf{x}_{0}$ to $\mathbf{x}_{1}$. As shown in the figure, we were able to add or remove concepts as we scaled up and down the related prompts.

\textbf{Comparison with prompt-to-prompt~\cite{p2p}.}We compared our approach with prompt-to-prompt on MM-CelebeA-HQ, as shown in Figure~\ref{fig:uspace_lpips}. To ensure fairness, we pre-trained a similar latent flow matching network, with the only difference being the use of U-VIT and UNet, respectively. We tuned the parameters of the UNet by hidden dimension to make it approximately equivalent to U-VIT. We kept the training strategy of learning rate and iteration number the same. During evaluation, we considered reweighting several prompts, such as the adjective before \textit{hair}, \textit{face}, \textit{noise}, and \textit{cheeks}. We conducted a total of 200 prompts and measure the fidelity to the unedited image using LPIPS perceptual distance score~\cite{zhang2018unreasonable_lpips} under different reweighting scales. Our method consistently outperformed prompt-to-prompt. We hypothesize that this is due to the fact that U-VIT is composed only of cross-attention, while UNet can apply attention only in the very intermediate layer for memory saving.
We would like to emphasize that in the visualization of Figure~\ref{fig:fm_t2i_vit_sequence} and ~\ref{fig:fm_t2i_vit_steering}, the background may appear slightly different due to changes in our target prompt. Although our transformer structure is attention-based, the encoder and decoder architecture in the latent flow matching framework includes convolution block, which may explain the slight background changes.

\textbf{Sampling Time analysis.} We maintain consistent complexity for both image editing and image sampling. Importantly, our semantic direction interpolation offers a versatile editing approach similar to samplings.

 \textbf{Failure cases.} We recognize that our method is not flawless, and there exist typical failure cases, such as over-constraining. For a more comprehensive understanding of these scenarios, we encourage readers to refer to the Appendix.

    \section{Related Work}

There has been a significant body of research dedicated to exploring the latent space of generative adversarial networks (GANs)~\cite{gandissection,shen2020interfacegan,shen2020interpreting}, which has provided valuable insights into enabling semantic editing of images through vector arithmetic in the latent space. 
Some recent works have extended similar explorations to diffusion models.
Kwon~\etal~\cite{hspace_kwon2022diffusion} have made an interesting discovery, identifying the bottleneck of the UNet denoising model as a suitable space for semantic image manipulation in diffusion models.
The discovery led to further exploration of the so-called $h$-space, resulting in supervised and unsupervised methods for discovering global semantic directions~\cite{park2023unsupervised, haas2023discovering}, image-specific semantic directions~\cite{haas2023discovering}, and style-aware semantic directions~\cite{jeong2023training},
In a different approach, Huberman-Spiegelglas~\etal~\cite{huberman2023edit} propose an alternative latent noise space that deviates from the traditional Gaussian distribution. This non-Gaussian noise space exhibits improved amenability to semantic image editing.
Our method differs in multiple aspects. 
In contrast to the aforementioned works, our method diverges in several key aspects. Firstly, we explore the semantic space of the U-ViT model~\cite{uvit} instead of the UNet. The U-ViT model has demonstrated favorable scaling behavior, but its dissimilarities from the UNet raise uncertainties and challenges in applying previous semantic editing methods that primarily rely on the spatially reduced latent representation at the bottleneck of the UNet.
Secondly, our focus lies on the general continuous normalizing flow (CNF) setting, allowing the utilization of generic ODE solvers that include adaptive step-size techniques. In contrast, previous editing methods were limited to fixed step-size solvers, rendering them incompatible with the efficiency benefits of adaptive step-size solvers.
%
%





Specialized for text-to-image diffusion models, prompt-to-prompt~\cite{p2p} leverages text-conditioned diffusion models for image editing by directly modifying the associated text prompt to achieve the desired semantic changes. 
The null-text inversion method~\cite{mokady2022null} extends this technique to real images by improving the inversion performance in the text-conditional setting.
%
%
DiffusionClip~\cite{kim2022diffusionclip} proposes to fine-tune a dedicated model for every type of editing, while DiffFit~\cite{difffit} enhances the efficiency of this by only fine-tuning a small subset of parameters.
It is important to note that these methods are primarily designed for UNet architectures, where text conditioning is incorporated through cross-attention mechanisms. In contrast, our exploration, focuses on U-ViT~\cite{uvit} architectures, revealing that the conclusions drawn from UNets may not necessarily apply to other architectural choices, such as U-ViTs. 
Furthermore, their method requires recasting the diffusion model as an ODE during inference, while Flow Matching enables direct learning of the ODE that we use during editing.
    \section{Conclusion}
In this paper, we explore the problem of image editing under the regime of Flow Matching with a transformer backbone. We discovered the $u$-space in this setting, which allows for editing in a controllable, accumulative, and composable manner. Furthermore, we make the editing method agnostic to the ODE solvers' choice of step size. To this end, we leverage the full-attention design in ViT and propose a simple and effective local-prompt method for augmenting the importance of specific parts of the prompt. As a result, this method can replace, remove, add, and scale the designated semantic concept. In future work, we plan to perform in-depth dissections to determine whether these conclusions hold true for other generative models. 

\clearpage
\bibliography{ref}

\clearpage


\onecolumn

\section{Supplementary Files}

\begin{figure*}
  \begin{center}
    \includegraphics[width=0.9\textwidth]{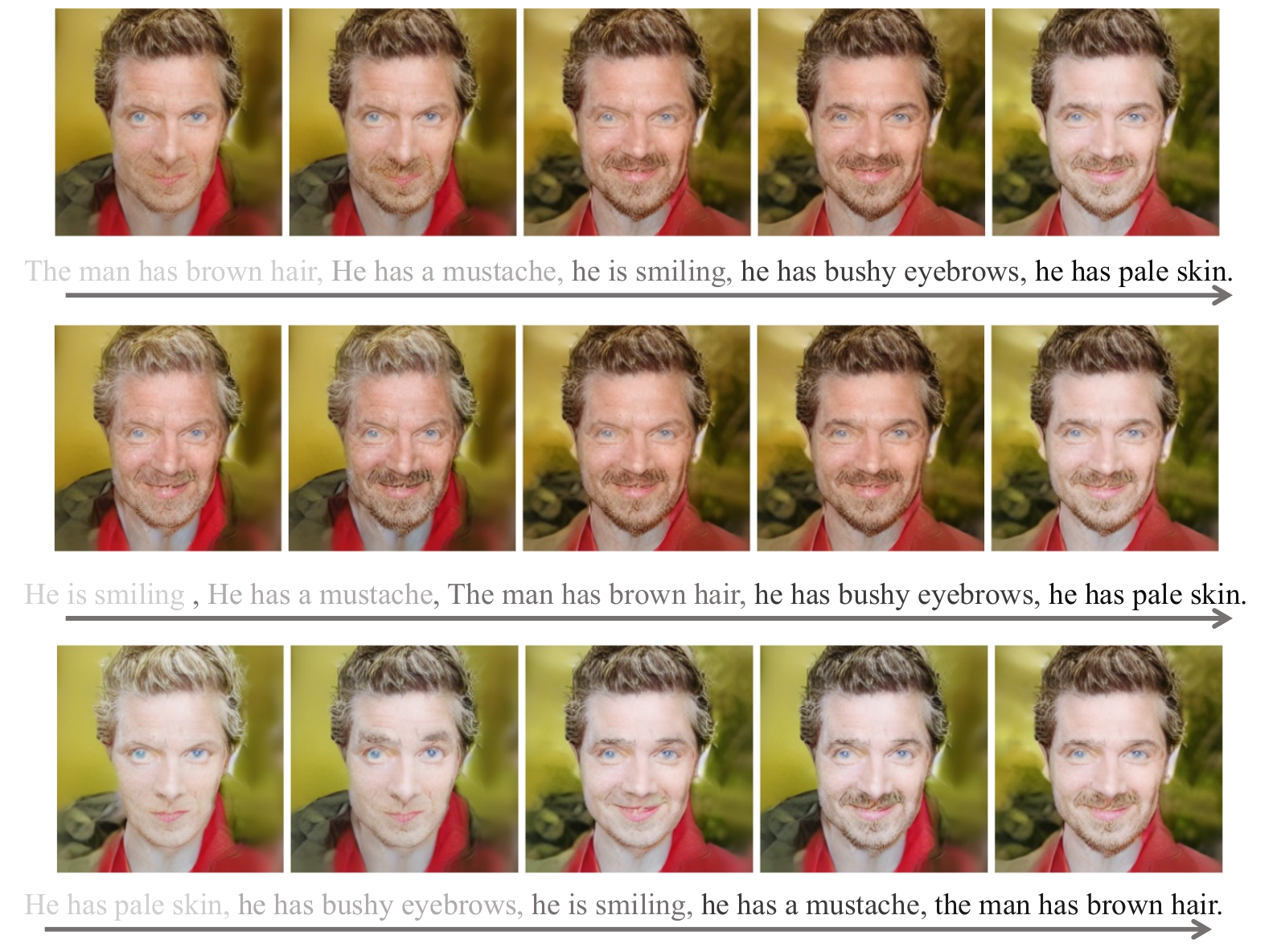}
  \end{center}
  \caption{\textbf{More sequential editing result by shuffling the order of the prompts.} The \textit{same} set of semantic manipulations can be composed in different orders.}
 \label{fig:fm_t2i_vit_sequence_appendix}
\end{figure*}

\begin{figure*}
    \centering
    \includegraphics[width=1.0\textwidth]{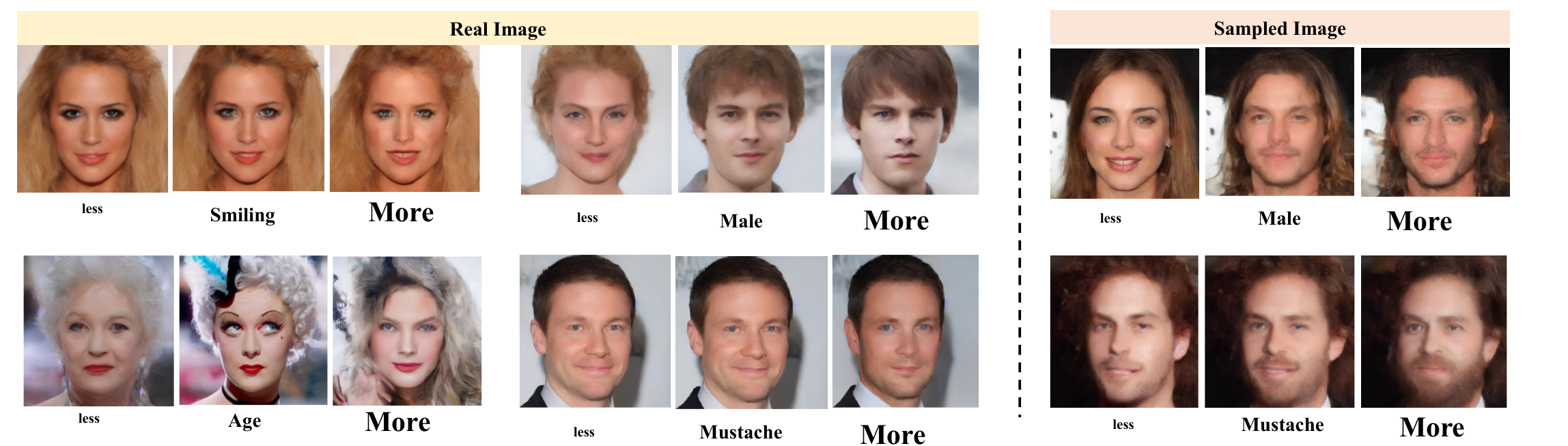}
    \caption{\textbf{Attribute editing on real images and sampled images.}}
    \label{fig:fm_manipulate_demo}
\end{figure*}

We further explore sequential editing in Figure~\ref{fig:fm_t2i_vit_sequence_appendix}. 
We shuffle the order of the prompts and find that the generated faces align with the corresponding prompt irrespective of the specific order. 
Interestingly, the final editing results (last column) are almost identical, indicating that the semantic edits can be composed through different prompt orders without altering the image layout.

\begin{algorithm}
        \caption{Semantic direction manipulation in $u$-space.}
 \label{alg:attribute_editing}
\begin{algorithmic}[1]
\STATE{\textbf{Input:} parameter of pretrained network $\theta$, fix step ODE solver $\text{ODE}_{f}$ with step N, adaptive ODE solver $\text{ODE}_{a}$, target attribute $k$.}
\STATE{\textbf{Output:} An edited image $\mathbf{x}_{1}$.} 

  \STATE{$\mathbf{x}_{0} \sim \mathcal{N}(0,I)$ a unit Gaussian random variable;}
 \STATE{$m = 0$;}
  \STATE{\textcolor{blue}{\texttt{\# Step 1. Collect the semantic direction in every timestep}}}
  \FOR{$i=1,2,\ldots,M$}
  \FOR{$j=N,N-1,\ldots,1$}
   \STATE{ $\mathbf{u}^{i}_{t_{j-1}},\mathbf{x}_{t_{j-1}} = \text{ODE}_{f} (\mathbf{x}_{t_{j}};\theta); \text{where } 
   t_{j}=\frac{j}{N} $ }
  \ENDFOR
 \ENDFOR
 \STATE{\textcolor{blue}{\texttt{\# Step 2. Calculate semantic direction}}}
\FOR{$j = N$ \TO $0$}
 \STATE{Calculate $\mathbf{s}_{t_{j}}^{k}$ by~\Cref{eq:linear_direction}} \COMMENT{Sample the index within the group}
\ENDFOR

 \STATE{\textcolor{blue}{\texttt{\# Step 3. Editing in backward process}}}
 
\WHILE{$t_{m} \le 1$}
 \IF{$0 < t < \tedit$}
 \STATE{$\text{Interpolate semantic direction }  \mathbf{s}^{k}_{t} \text{ by ~\Cref{eq:interp}}$;}
 \STATE{$\mathbf{x}_{t_{m+1}}, t_{m+1} = \text {ODE}_{a}(\mathbf{x}_{t_{m}}, \mathbf{u}_{t_{m}} + w \cdot \mathbf{s}^{k}_{t_{m}}, t_{m};\theta)$;}
 \ELSE
 \STATE{$\mathbf{x}_{t_{m+1}} , t_{m+1}= \text {ODE}_{a}(\mathbf{x}_{t_{m}}, \mathbf{u}_{t_{m}},  t_{m}; \theta)$;}
 \ENDIF
 \ENDWHILE
\STATE {\textbf{Return} $\textbf{x}_{1}$}

\end{algorithmic}
\end{algorithm}

\subsection{Hyperparameters choices and details}

We perform experiments on three datasets and document our hyperparameter choices in Table~\ref{tab:setting}. 
For sampling, we utilize the torchdiffeq~\cite{neuralode_chen2018neural} library  with \texttt{atol=rtol=1e-5}.

We generate the initial image with a fixed seed and a given initial prompt. Then we concatenate additional captions to the prompt one by one and sequentially edit the image.
Currently, we only support editing for images that have a caption. We leave the editing of images without captions to future work.



\begin{table*}
\begin{center}
\begin{tabular}{lcccccc}
\toprule
    Dataset  & CelebA 256$\times$256 & MM-CelebA-HQ 256$\times$256 & MS-COCO 256$\times$256 \\
    \midrule
    Latent shape  & 32$\times$32$\times$4 & 32$\times$32$\times$4 & 32$\times$32$\times$4 \\
    U-ViT type & U-ViT-L/2& U-ViT-L/2& U-ViT-L/2\\
    depth & 20 & 20 & 20 \\
    embedding dim & 1024 & 1024 & 1024 \\
    num of head & 16  & 16 & 16\\
    \midrule
    Batch size & 256 & 256 & 512\\
    Training iterations & 100K & 100K & 200k \\
    GPU & 4 $\times$ A5000 & 4 $\times$ A5000 &4 $\times$ A5000\\
    \midrule
    Optimizer & Adam & Adam & Adam\\
    Learning rate & 1e-4 & 1e-4 & 1e-4 & \\
    \midrule
    $\sigma_{min}$  &1e-4 & 1e-4 & 1e-4 \\
    \bottomrule
\end{tabular}

\end{center}
\caption{\textbf{Hyperparameter Settings} of three datasets.}
\label{tab:setting}
\end{table*}

\subsection{More related works}

\paragraph{Cross-attention in generative models.}

Cross-attention is broadly applied in generative model to incorporate the guidance signal e.g., text~\cite{rombach2022high_latentdiffusion_ldm,saharia2022photorealistic_imagen}, bounding box~\cite{sgdm}, segmentation mask~\cite{rombach2022high_latentdiffusion_ldm}. Prompt-to-Prompt~\cite{p2p} explores the cross-attention between the text and visual modalities to achieve controllable generation. Instead of fine-tuning based on CLIP difference in DiffusionClip~\cite{kim2022diffusionclip},~\cite{forgetmenot} utilize QKV attention Re-steering to achieve concept negation directly on pre-trained models.~\cite{difffit} only fine-tune attention's bias weight to conduct parameter-efficient fine-tuning to achieve transferability on other datasets.~\cite{rin_jabri2022scalable} utilize cross-attention to exchange information between latent tokens and image tokens. On the side hand, ViT~\cite{dosovitskiy2020image_vit} has been applied as a surrogate of UNet~\cite{unet} in diffusion models, e.g., U-ViT~\cite{uvit}, DiT~\cite{dit_peebles2022scalable}, GenViT~\cite{genvit},RIN~\cite{rin_jabri2022scalable}.
We focus on the exploration of the U-ViT structure in flow matching, and the U-ViT composed of full attention blocks has not been explored before.

\subsection{Visual comparison between different ODE solvers for the semantic direction interpolation}

\begin{figure}
  \begin{center}
    \includegraphics[width=0.5\textwidth]{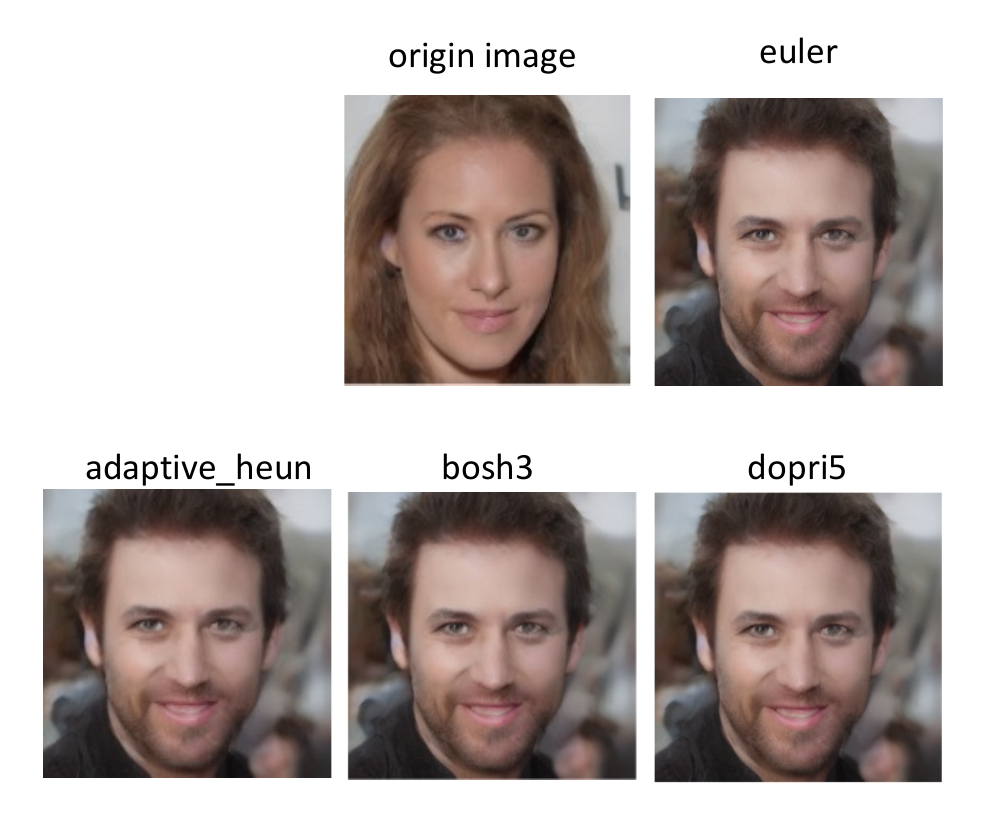}
  \end{center}
  \caption{\textbf{The editing visualization by semantic direction interpolation.} The default ODE solver is \texttt{euler} with 100 time steps. We can find that interpolation indeed leads to visual similar manipulation.}
 \label{fig:vector_field_interp}
\end{figure}

In~\Cref{fig:vector_field_interp}, we invert the original image by solving the ODE in the forward direction using an \texttt{euler} solver for 100 steps. Then, we change the \textit{gender} attribute by manipulating the latent noise with a weight of $w=1$. Afterwards, we compare different ODE adaptive step-size solvers with the fixed step-size \texttt{euler} solver. We observe that the adaptive step size solvers (\texttt{dopri5},~\texttt{bosh3},~\texttt{adaptive heun}) achieve visually very similar results when compared to the more expensive \texttt{euler} method. 

\subsection{PCA in $u$-space}

\begin{figure*}
  \begin{center}
    \includegraphics[width=0.9\textwidth]{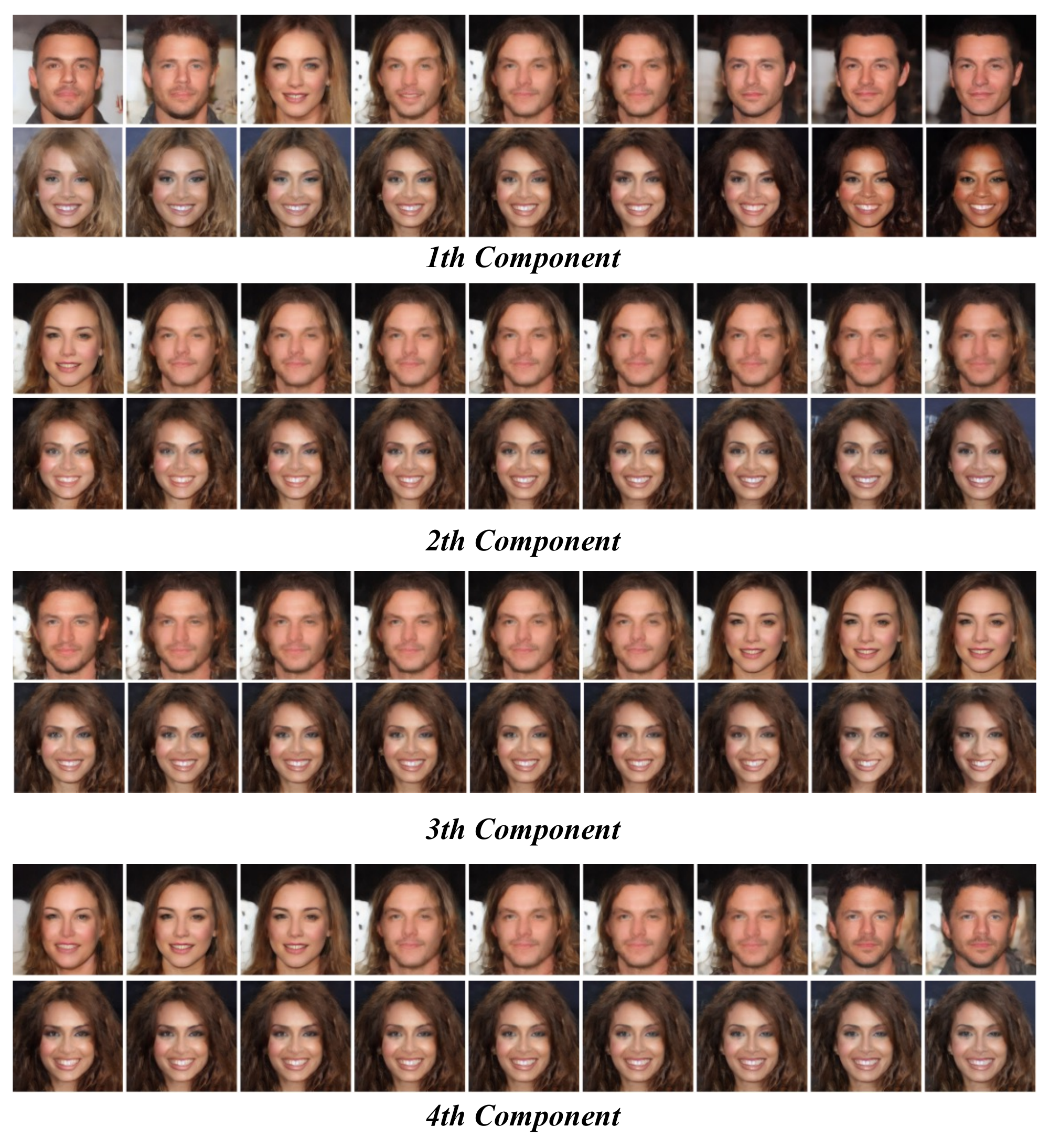}
  \end{center}
  \caption{\textbf{Semantic direction from the top 4 components of PCA.} We generate manipulated samples of two identical images by adjusting the guidance strength range to [-2, -1.5, -1, -0.5, 0, +0.5, +1, +1.5, +2]. These manipulated samples demonstrate that the directions found by PCA do not yield consistent semantic manipulations.}
 \label{fig:pca_is_noisy}
\end{figure*}

In detail, we perform PCA on 10,000 samples with shape $4 \times 32 \times 32$ in $u$-space. The two-dimensional features $\mathbf{u}(\mathbf{x}_{i,t})$ from the U-ViT are flattened into one dimensional vectors before applying PCA. We show some sample augmentations along the first four principal components in~\Cref{fig:pca_is_noisy}. We observe that for different samples there is a difference in the type of semantic manipulation that occurs. We hypothesize that the lack of spatial compression in U-ViT leads to this phenomenon, where a linear augmentation along the principle components can augment multiple different semantic aspects of the image.

\subsection{Semantic editing in other blocks of U-ViT}

We ablate the effectiveness of the intermediate layer for semantic manipulation. The tensor shape is [T, C], where $T=257$ and $C=1024$. We find that the tensor is too large to calculate the semantic direction, which can lead to unstable results. In total, the overall dimensions are 64 times larger than $u$-space (\eg $257 \times 1024 \approx   64 \times (4 \times 32 \times 32) $).

Based on the above analysis, we choose to perform semantic editing at the beginning of U-ViT, before the patchifying and attention blocks.

\subsection{Attention map visualization in Local-Prompt}

\begin{figure*}
  \begin{center}
    \includegraphics[width=0.99\textwidth]{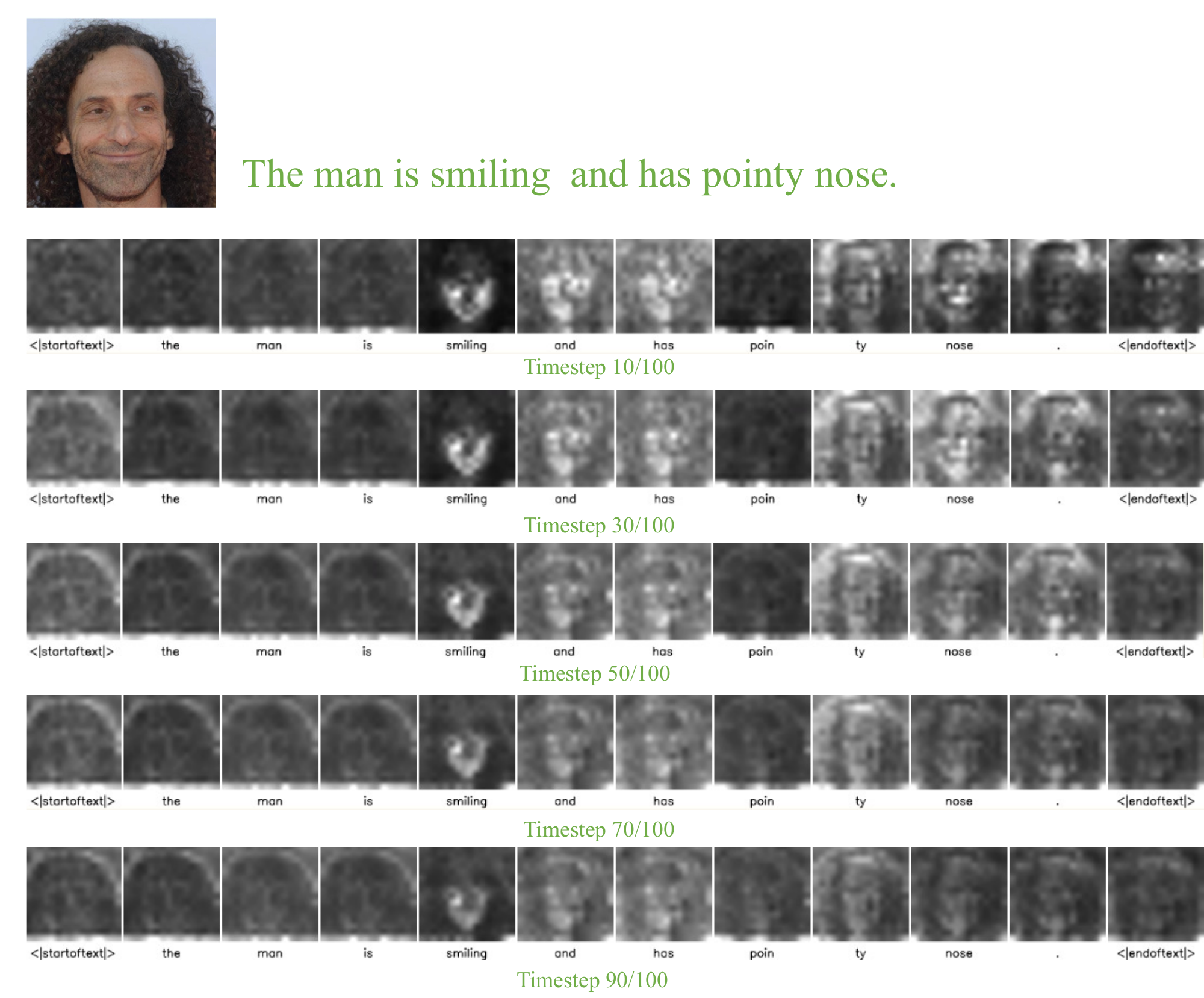}
  \end{center}
  \caption{\textbf{Attention map of different timestep of \textit{3th}-Block in U-VIT.} }
 \label{fig:attention_map_vis}
\end{figure*}
\begin{figure*}
  \begin{center}
    \includegraphics[width=0.99\textwidth]{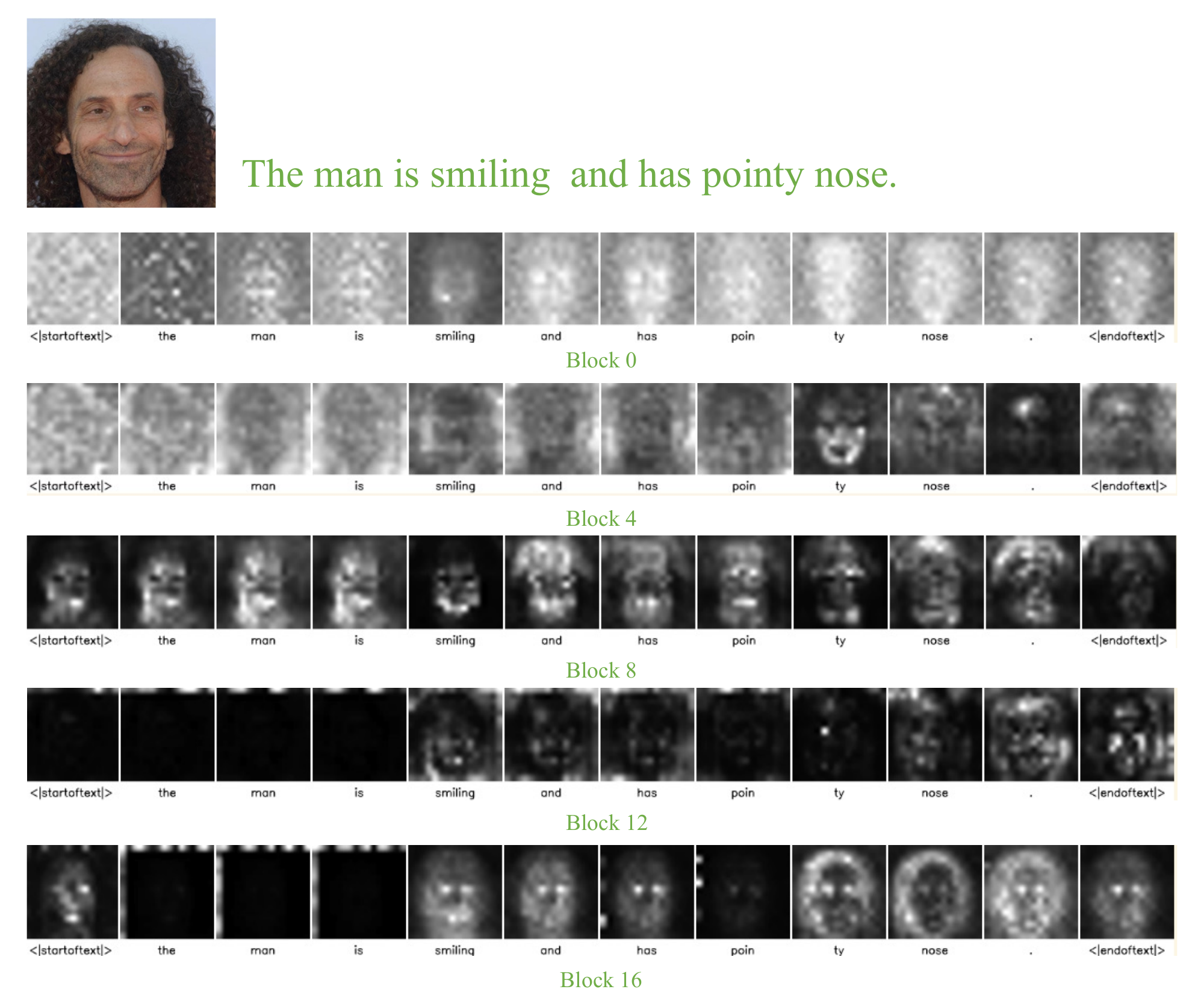}
  \end{center}
  \caption{\textbf{Attention map of different blocks of \textit{10th} step in 100 timesteps in U-ViT.} }
 \label{fig:attention_map_vis_blocks}
\end{figure*}

We visualize the attention map token by token and normalize the image value by image. We illustrate the attention map of different blocks in \Cref{fig:attention_map_vis_blocks} and the attention map of different time steps in \Cref{fig:attention_map_vis}. We find that early timesteps demonstrate attention maps that clearly outline specific parts of the face in comparison to later time steps. This observation aligns well with the conclusion presented in the main paper. In particular, at earlier stages during the sampling process, the transformer requires the text prompt to specify the to-be-generated content of the image. At later steps the self-attention can attend to other regions in the image instead of the text prompt. Secondly, we observe that the highest activation is achieved in \textit{block 0} and that the most discriminative blocks are \textit{block 4} and \textit{block 8}.

\subsection{Prompt rescaling in every block}

As we saw in~\Cref{fig:attention_map_vis_blocks}, we observe that different blocks exhibit different magnitudes for the attention values. This highlights the significance of the magnitude of the attention and motivates us to perform attention-rescaling for the whole block.

\rot{Why does local-prompt editing work? and comparison with prompt-to-prompt?}
\rot{Local-prompt on COCO dataset?}

\subsection{Exploration in early time steps}

\begin{figure*}
    \centering
    \includegraphics[width=0.8\textwidth]{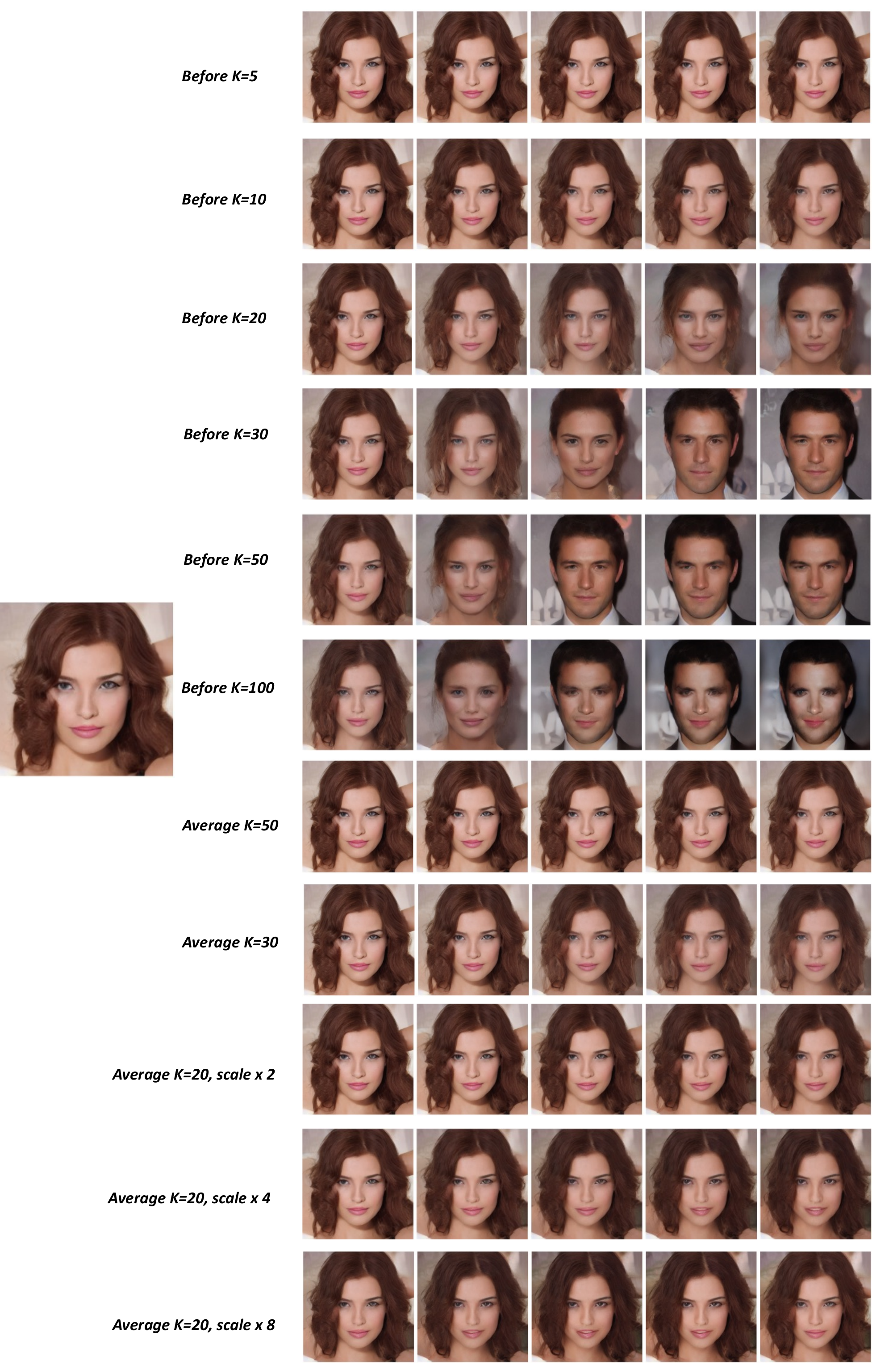}
    \caption{\textbf{Early timestep has semantic editing ability.} The default scale $w$ is [0.5, 1, 1.5, 2, 2.5]. The edited attribute is \textit{male}.}
    \label{fig:early_time_has_semantic}
\end{figure*}

By visualizing the progressive sampling procedure, we explore how incorporating the augmentation at different timesteps during sampling affects the editing process.
Our results show that incorporating the augmentations in the early timesteps works well, while evenly splitting the same number of manipulation operations throughout the entire period is not effective in manipulating the achieving the desired editing.
Additionally, we investigate the effect of increasing the weight value $w$ and find that this can disturb the final sample quality and discard the content of the source image.
These findings are presented in \Cref{fig:early_time_has_semantic}. Our study provides insights into effective techniques for semantic manipulation and highlights the importance of early timestep guidance in this process.

\subsection{More visualization}

\begin{figure*}
  \begin{center}
    \includegraphics[width=1.0\textwidth]{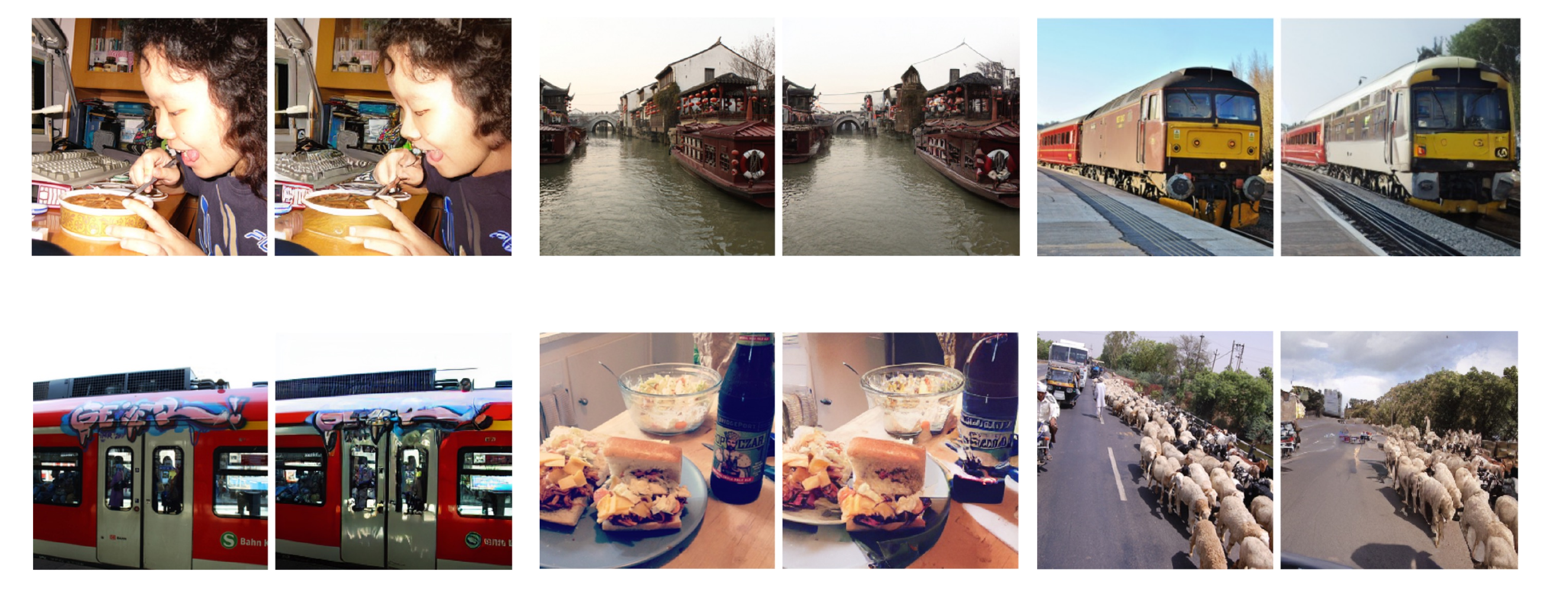}
  \end{center}
  \vspace{-10pt}
  \caption{\textbf{Adding noise prompts does not significantly change the layout.} However, please note that the content may be slightly altered due to the marginal semantic relationship between noise prompts and visual tokens.}
 \label{fig:adding_noise_prompt_layout_doesnt_change}
\end{figure*}

We add a noise prompt to the MS COCO dataset and find that it did not significantly alter the layout, as shown in~\Cref{fig:adding_noise_prompt_layout_doesnt_change}.

\begin{figure*}
    \centering
    \includegraphics[width=1.0\textwidth]{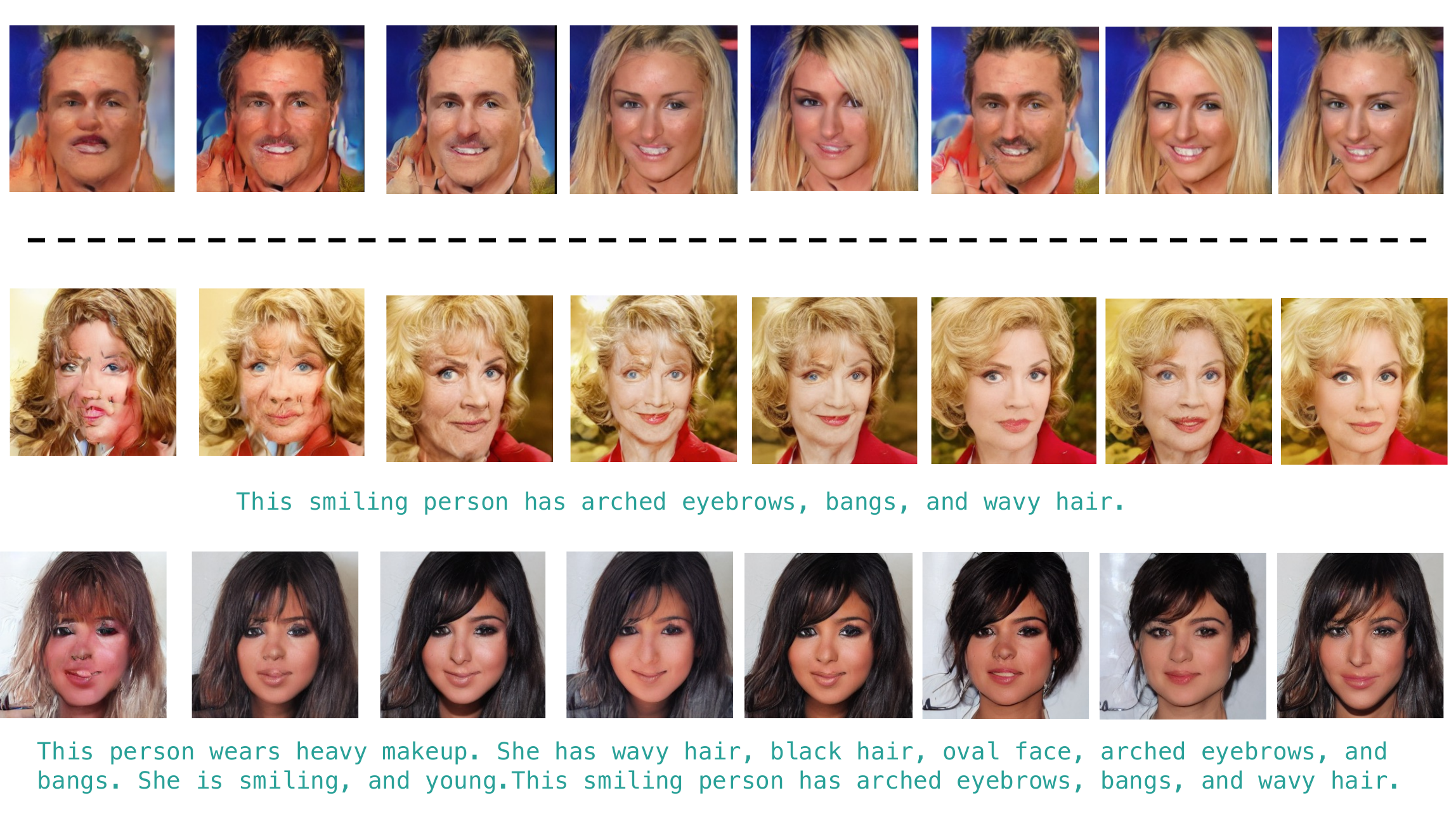}
    \vspace{-15pt}
    \caption{\textbf{Training samples with the same latent noise and same prompt.} The first row contains examples of image generation, while the second and third rows contain examples of text-conditioned image generation. Best viewed in color.}
    \label{fig:train_progress_vis}
\end{figure*}

We visualize the samples with the same latent noise and prompt during the training process in \Cref{fig:train_progress_vis}.

\subsection{Failure Case}

\begin{figure*}
    \centering
    \includegraphics[width=1.0\textwidth]{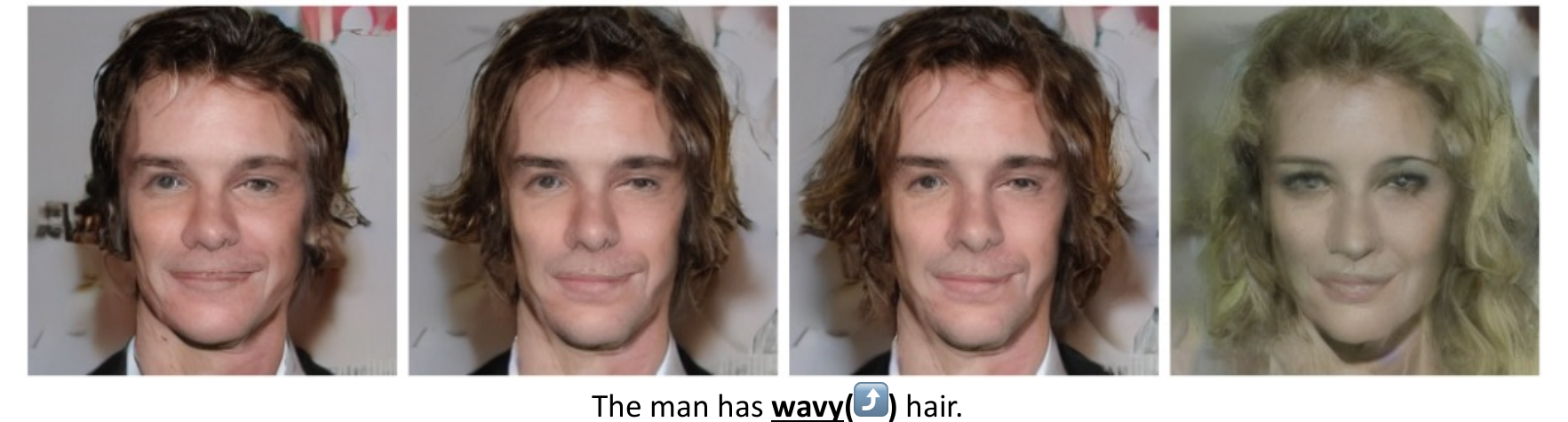}
    \caption{\textbf{Failure Case: too large values for the self-attention rescaling will lead to over-constraining.} The \textit{wavy} attribute can cause the generated image to be \textit{female} if the scaling factor is too large, e.g., $s=25$.}
    \label{fig:self_attention_over_constrain}
\end{figure*}

If the scaling factor $s$ during self-attention is too large, it can lead to over-constraining. This is demonstrated in~\Cref{fig:self_attention_over_constrain}.

\begin{figure*}
    \centering
    \includegraphics[width=1.0\textwidth]{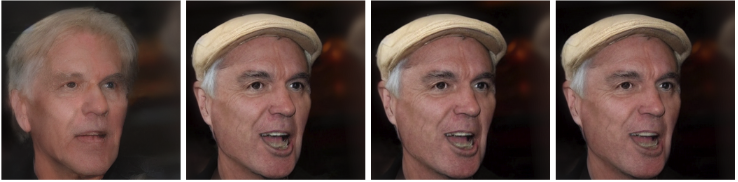}
    \caption{\textbf{Failure Case: scaling up the \textit{male} attributes on a male image will lead to concept entanglement.} This is because increasing the \textit{male} attribute can cause the generated image to include a \textit{hat}, which is not necessarily a male attribute.}
    \label{fig:failurecase2}
\end{figure*}

If we add the \textit{male} attribute to a male image, it can lead to concept entanglement, as demonstrated in~\Cref{fig:failurecase2}.

\end{document}